\documentclass{article}

\usepackage[nonatbib]{neurips_2022}

\usepackage[utf8]{inputenc} 
\usepackage[T1]{fontenc}    
\usepackage{hyperref}       
\usepackage{url}            
\usepackage{booktabs}       
\usepackage{amsfonts}       
\usepackage{nicefrac}       
\usepackage{microtype}      
\usepackage{xcolor}         
\usepackage[numbers]{natbib}
\usepackage{graphicx}

\graphicspath{{fig/}}

\title{Privacy Preserving Image Classification - Final Report}

\author{%
  Basanta Chaulagain\\
  Department of Computer Science\\
  University of Georgia\\
  Athens, GA \\
  \texttt{basanta.chaulagain@uga.edu} \\
}

\begin{document}
\nolinenumbers

\maketitle

\begin{abstract}
  This report details the final results of our project in CSCI 8960: Privacy-Preserving Data Analysis. The project aims to develop differentially private deep learning models for image classification on CIFAR-10 datasets \cite{cifar10} and analyze the impact of various privacy parameters on model accuracy. We have implemented five different deep learning models, namely ConvNet, ResNet18, EfficientNet, ViT, and DenseNet121 and three supervised classifiers namely K-Nearest Neighbors, Naive Bayes Classifier and Support Vector Machine. We evaluated the performance of these models under varying settings. Our best performing model to date is EfficientNet with test accuracy of $59.63\%$ with the following parameters (Adam optimizer, batch size 256, epoch size 100, epsilon value 5.0, learning rate $1e-3$, clipping threshold 1.0, and noise multiplier 0.912).
\end{abstract}

\section{Introduction}

In recent years, deep learning models have shown remarkable success in a wide range of applications, including image classification, natural language processing, and speech recognition \cite{obaid2020deep}. However, the increasing concern over privacy has raised questions about the use of sensitive data in these applications. To address these concerns, differential privacy (DP) has emerged as a popular approach for privacy-preserving data analysis. 

Differential privacy is a mathematical framework for ensuring the privacy of individuals in datasets. It can provide a strong guarantee of privacy by allowing data to be analyzed without revealing sensitive information about any individual in the dataset \cite{devaux2022}. The core idea behind DP is to add controlled noise to the computation so that the output is indistinguishable with high probability, whether or not a particular individual's data is included in the computation.

In this project, we explore the application of DP to deep learning models for image classification on the CIFAR-10 dataset \cite{cifar10}. We evaluate the performance of five different deep learning models - ConvNet, ResNet18, EfficientNet, ViT and DenseNet121 and three supervised classfiers - K-Nearest Neighbors, Naive Bayes Classifier and Support Vector Machine - under various privacy settings.

We first provide a brief overview of the models we selected and the rationale behind our choices. We then describe the architecture of each model, any modifications we made to the original architecture, and related work on the use of these models in image classification. We also discuss the training details of the models, including data pre-processing, training objectives, approach to satisfying $(\epsilon, \delta)-DP$, optimizer details with hyperparameter values, and DP details such as budget accounting, noise multiplier, and number of epochs.

We then present the results of our experiments, including train and test loss and accuracy against epochs and an ablation study to understand the effects of hyperparameters on the models' accuracy. Finally, we discuss the limitations of our work and future directions for improving the accuracy of our models.

\section{Related work}

In recent years, the privacy-preserving deep learning techniques have gained popularity due to the increasing concerns over privacy while handling sensitive data in various applications. Researchers have proposed different privacy-preserving techniques to address this challenge, and various works have been conducted to apply these techniques to image classification tasks.

Shokri et al. \cite{shokri2015privacy} proposed a differentially private algorithm that uses the moment accountant to train deep neural networks. Abadi et al. \cite{abadi2016deep} proposed TensorFlow Privacy, which is a deep learning framework that implements DP for training deep neural networks. Zhao et al. introduced the Private Aggregation of Teacher Ensembles (PATE) \cite{zhao2021privacy} method that enables the training of differentially private models without access to the raw training data. Huang et al. proposed DPDenseNet \cite{Huang2017Densely}, which uses a dense block structure to improve the accuracy of differentially private image classification.

Some researchers have investigated transfer learning techniques in differential privacy's hybrid model, as presented in Kohen et al. \cite{kohen2022transfer}. Other works have focused on reconstructing training data while maintaining privacy, as in Guo et al. \cite{guo2022bounding}, and designing efficient private inference techniques, as in Cho et al. \cite{cho2022selective}.

Moreover, several recent works have explored different trade-offs between privacy and model performance. Mireshghallah et al. \cite{mireshghallah2022differentially} proposed differentially private model compression, while Bu et al. \cite{bu2022scalable} focused on scalable and efficient training of large convolutional neural networks with differential privacy. Tan et al. \cite{tan2019efficientnet} presented a provable tradeoff between overparameterization and membership inference. Wang et al. \cite{wang2022renyi} introduced Renyi Differential Privacy of Propose-Test-Release and its applications to private and robust machine learning.

In terms of deep learning models for image classification, Convolutional Neural Networks (ConvNets) are the most commonly used. Krizhevsky et al. introduced AlexNet \cite{krizhevsky2012imagenet}, a deep ConvNet that achieved state-of-the-art accuracy on the ImageNet dataset. Since then, several variations of ConvNets have been proposed, such as ResNet \cite{he2016deep}, DenseNet \cite{Huang2017Densely}, EfficientNet \cite{tan2019efficientnet}, and Vision Transformer (ViT) \cite{dosovitskiy2021image}. These models have shown to achieve even better accuracy on various image classification tasks.

\section{Classificaiton Models}
For this project, we have selected five popular deep learning models: ConvNet, ResNet18, EfficientNet, ViT and DenseNet121, and three supervised classifiers: SVM, NBC and KNN. Our rationale for choosing these models is to cover a broad range of popular deep learning architectures that have been widely used in various computer vision tasks.

\subsection{ConvNet}
ConvNet is one of the simplest and widely used CNN-based models for classifying image dataset. We chose ConvNet because it is one of the simplest and widely used CNN-based models for image classification, which will help us gain a basic understanding of CNNs and their application in differential privacy settings. Recent works such as Guo et al. \cite{guo2022bounding} have implemented ConvNets under differential privacy settings for image classification tasks.

\subsection{ResNet18}
ResNet18 is a deeper version of the ResNet family of models and has been widely used in various image classification tasks. We chose ResNet18 to evaluate its performance under differential privacy settings and compare it with other models. Recent works, such as Kohen and Sheffet \cite{kohen2022transfer} and Mireshghallah et al. \cite{mireshghallah2022differentially}, have implemented ResNet models under differential privacy settings for image classification and compression tasks. Additionally, Alkhelaiwi et. al. used this model \cite{alkhelaiwi2021efficient} in their work to classify satellite images.

\subsection{EfficientNet}
EfficientNet is a CNN-based model that provides one of the state-of-the-art accuracy for image classification with efficient resource usage under non-private setting. We chose EfficientNet to evaluate its performance under differential privacy settings and compare it with other models. Recent works such as Tan et. al. \cite{tan2019efficientnet} have implemented EfficientNet under differential privacy settings for training large convolutional neural networks.

\subsection{ViT}
ViT is a transformer-based model, which replaces the convolutional layers of traditional CNNs with self-attention layers, that enable the model to capture global image features by attending to all positions in the input image. We chose ViT to evaluate its performance under differential privacy settings and compare it with other models. Recent works such as Tan et al. \cite{tan2019efficientnet} have analyzed the privacy-accuracy tradeoff for ViT models under differential privacy settings.

\subsection{DenseNet121}
DenseNet121 is a popular CNN-based model that has been widely used in various image classification tasks. We chose DenseNet121 as it has shown to be effective in differential privacy settings as in the work \cite{alkhelaiwi2021efficient} done by Alkhelaiwi et. al. Recent works, such as Huang et al. \cite{Huang2017Densely}, have proposed DPDenseNet, which uses a dense block structure to improve the accuracy of differentially private image classification.

\subsection{Supervised Classifiers}
In addition to the above deep learning models, we also used supervised classifiers such as SVM, KNN, and NBC for our experiments. We wanted to see how additional noise affects the accuracy of these classical classifiers. We chose these models because they have been tried by other researchers in the context of privacy-preserving image classification tasks, such as Senekane et al. \cite{senekane2019} and Wang et al. \cite{wang2020privacy}.

SVM is a popular algorithm for classification and regression analysis, widely used in image classification tasks for handling high-dimensional data effectively. KNN is a simple and effective algorithm for classification and regression analysis, working by finding the k-nearest neighbors and classifying based on majority label. NBC is a probabilistic algorithm for classification, widely used in image classification due to its simplicity and efficiency. It calculates the probability of input belonging to a certain class based on training data.

\section{Evaluation}
We conducted experiments on the five deep learning models mentioned earlier, where we varied the privacy settings. In total, 70+ experiments were performed for different models by assigning different hyper-parameters values such as epsilon, batch size, clipping threshold, noise multiplier, optimizer, learning rate and epoch size. Table~\ref{experimental_setup} presents the details of the eight experiments for each model. The value of delta ($\delta$), was kept constant at $1e-5$ for all experiments. We couldn't compute the performance of ResNet18 and DenseNet121 for epoch size 256 due to the limitation of the resource in the machine.

We used the python libraries PyTorch \cite{pytorch} and Opacus \cite{opacus} for implementing differential privacy in deep learning models. For the training and testing environment, we used the GPUs from Google Colab \cite{googlecolab} and Kaggle notebook \cite{kaggle}. We used ResNet, pre-trained DenseNet, pre-trained EfficientNet and pre-trained ViT from the PyTorch library, while we constructed a simple ConvNet with four convolutional layer, and one linear layer.

\begin{table}
  \caption{Experimental Setup showing different values of epsilon, batch size, clipping threshold, noise multiplier, and optimizer.}
  \label{experimental_setup}
  \centering
  \begin{tabular}{p{0.5cm}p{1cm}p{1cm}p{0.75cm}p{1cm}p{1cm}p{0.75cm}p{1cm}p{2.5cm}p{1cm}}
    \toprule
     Exp.  & Optim     & Batch Size   & Epsilon    & Clipping threshold    & Learning rate  & Epochs    & Noise Multiplier  & Models    & \# of runs \\
    \midrule
        1	& SGD	& 128	& 20	& 1	& 1e-3	& 50	& 0.47	& M1, M2, M3, M4	& 1 \\
        2	& Adam	& 128	& 20	& 1	& 1e-3	& 50	& 0.47	& M1, M2, M3, M4	& 1 \\
        3	& Adam	& 128	& 5	& 1	& 1e-3	& 50	& 0.67	& M1, M2, M3, M4	& 1 \\
        4	& RMSProp	& 128	& 5	& 1	& 1e-3	& 50	& 0.67	& M1, M2, M3, M4	& 1 \\
        5	& Adam	& 128	& 5	& 0.75	& 1e-3	& 50	& 0.67	& M1, M2, M3, M4	& 1 \\
        6	& Adam	& 128	& 5	& 0.5	& 1e-3	& 50	& 0.67	& M1, M2, M3, M4	& 1 \\
        7	& Adam	& 128	& 2.5	& 0.75	& 1e-3	& 50	& 0.88	& M1, M2, M3, M4	& 1 \\
        8	& Adam	& 256	& 2.5	& 0.75	& 1e-3	& 50	& 1.07	& M1, M3, M4	& 1 \\
        9	& Adam	& 128	& 5	& 1	& 1e-3	& 100	& 0.76	& M1, M2, M3, M4, M5	& 3 \\
        10	& Adam	& 256	& 5	& 1	& 1e-3	& 100	& 0.91	& M1, M3, M4	& 3 \\
        11	& Adam	& 256	& 5	& 1	& 1e-2	& 100	& 0.91	& M1, M3, M4	& 3 \\
        12	& Adam	& 128	& 5	& 1	& 5e-4	& 100	& 0.76	& M1, M2, M3, M4	& 3 \\
        13	& Adam	& 256	& 5	& 1	& 5e-4	& 100	& 0.91	& M1, M3, M4	& 3 \\
        14	& Adagrad	& 256	& 5	& 1	& 1e-4	& 100	& 0.91	& M1, M3, M4	& 3 \\
        15	& Adam	& 256	& 5	& 1.5	& 5e-4	& 100	& 0.91	& M1, M3, M4	& 3 \\
        16	& Adam	& 256	& 3	& 2.5	& 5e-4	& 100	& 1.21	& M1, M3, M4	& 3 \\
        17	& Adam	& 256	& 5	& 5	& 5e-4	& 100	& 0.91	& M1, M3, M4	& 3 \\
        18	& Adam	& 256	& 3	& 5	& 5e-4	& 100	& 1.21	& M1, M3, M4	& 3 \\
        19	& Adam	& 256	& 1	& 5	& 5e-4	& 100	& 2.81	& M1, M3, M4	& 3 \\
        20	& Adam	& 128	& 5	& 1	& 1e-3	& 200	& 0.91	& M1, M2, M3, M4	& 1 \\
    \bottomrule
  \end{tabular}
\end{table}

In this study, we examine the effect of various parameters on the performance of the models. To evaluate model performance, we measure the test and train accuracy as well as the loss. In this report, we present the expectation of performance metrics for multiple runs of each experiment report their magnitude. Since we draw random samples for training, we calculate the expectation for each metrics. Due to the limitation of time and resource, we have performed at most 3 runs for each experiments, and calculated expectation to report the metrics.

\subsection{Epsilon}
\begin{figure}
    \centering
    \includegraphics[scale=0.5]{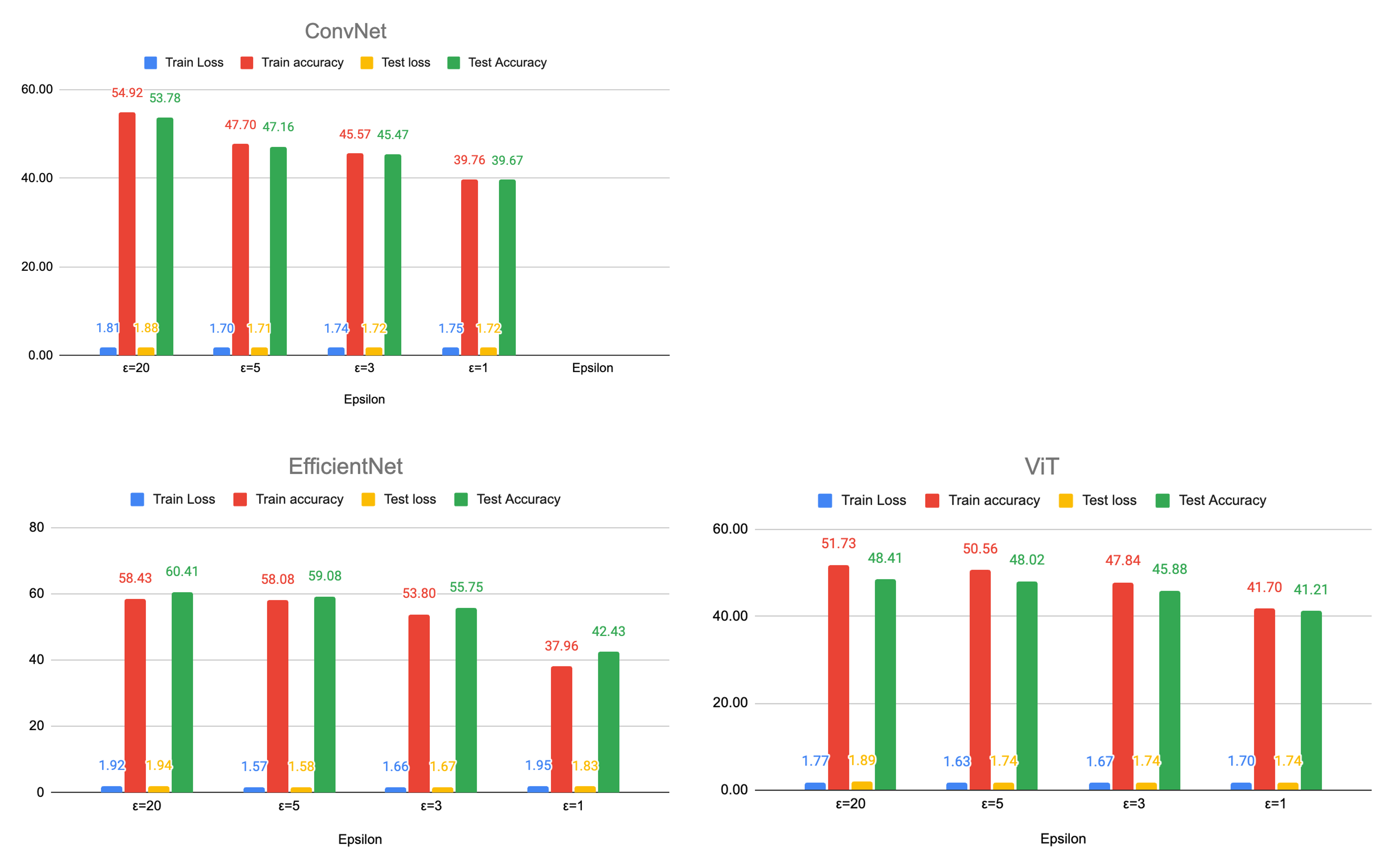}
    \caption{\small Performance of different models when $\epsilon$ = 20, 5, 3 and 1. All experiments implement Adam optimizer.}
    \label{fig:var_epsilon}
\end{figure}

We analyzed the effect of privacy parameter $\epsilon$ on the model accuracy using experiments 2, 17, 18 and 19. These experiments were conducted with Adam optimizer. We studied the performance for the epsilon value of $\epsilon$ = 20, 5 and 3 and 1 for models in experiments 2, 17, 18 and 19. The results indicated that as the value of $\epsilon$ decreases, the accuracy of all four models also decreases. The EfficientNet model showed the highest test accuracy for each values of $\epsilon$ in our experiments. The performance metrics for these models are shown in figure \ref{fig:var_epsilon}.

\subsection{Clipping threshold (C)}
\begin{figure}
    \centering
    \includegraphics[scale=0.5]{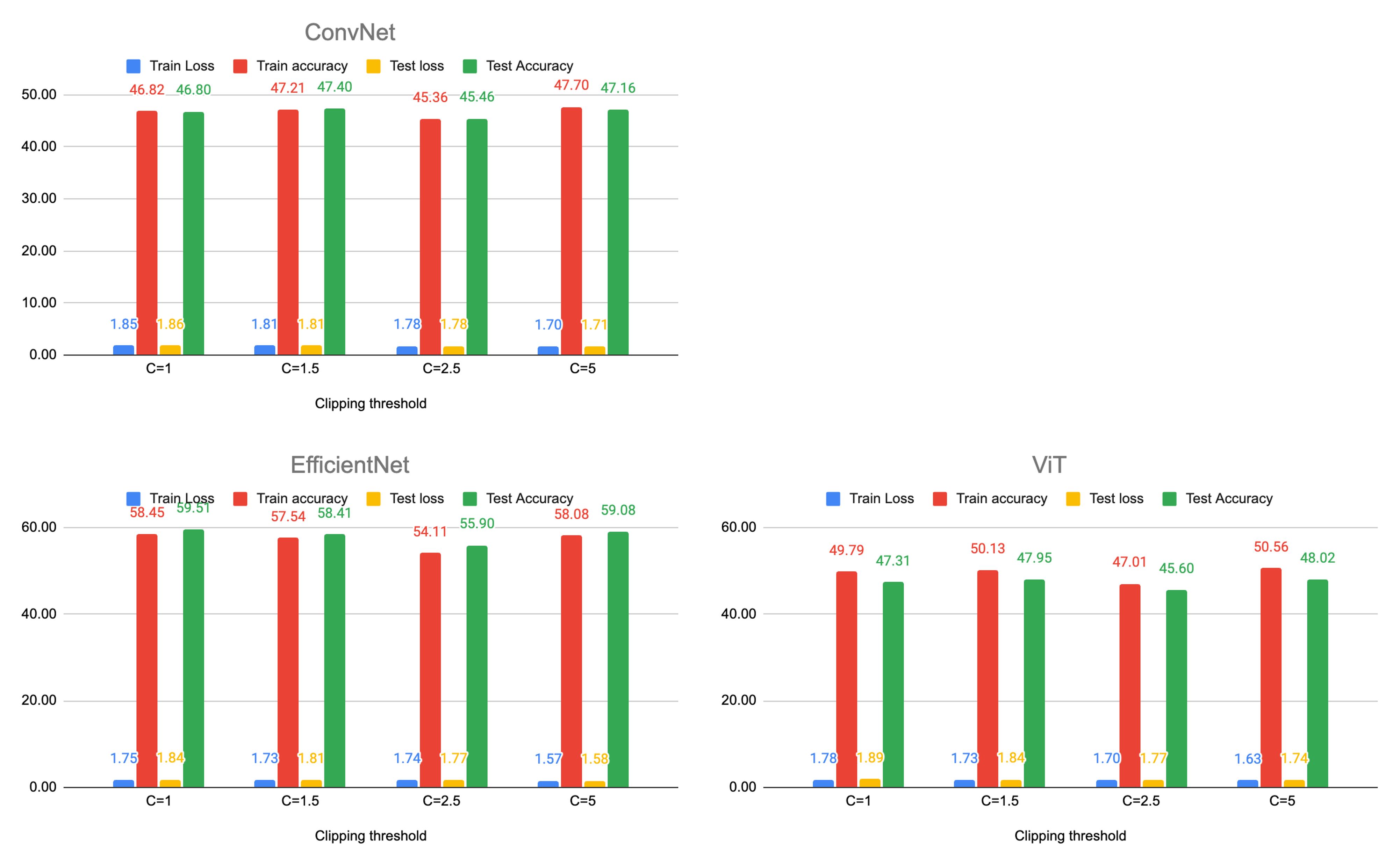}
    \caption{\small Performance of different models when $c$ = 1, 1.5 and 2.5 and 5. All experiments implement Adam optimizer, a batch size of 256, $\epsilon$ 5.0 and learning rate of 5e-4.}
    \label{fig:var_c}
\end{figure}

To understand the effect of clipping threshold ($C$) on model accuracy, we conducted experiments 3, 5, and 6 under one settings and 13, 15, 16 and 17 under other settings. We report the performance of later 4 experiments which uses  Adam optimizer, a batch size of 256, and $\epsilon$ value of 5.0 and learning rate of 5e-4. We studied the performance for the clipping threshold value of 1, 1.5, 2.5 and 5. Our results demonstrated that an increase in $C$ improved the model accuracy for all four models. The performance metrics for these models are shown in figure \ref{fig:var_c}.

\subsection{Optimizer}
\begin{figure}
    \centering
    \includegraphics[scale=0.5]{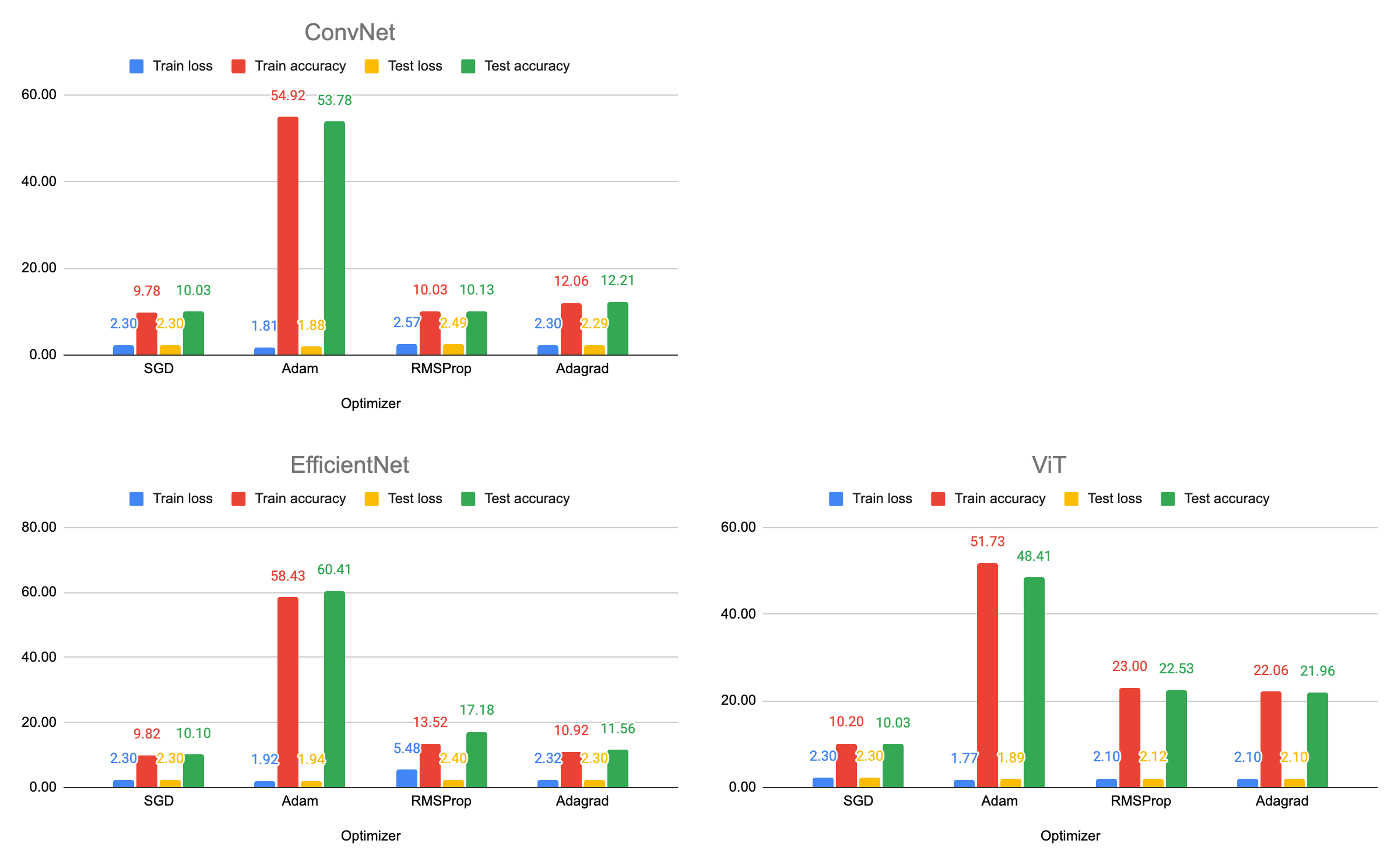}
    \caption{\small Performance of different models when the optimizer used are SGD, Adam, Adagrad and RMSProp.}
    \label{fig:var_optimizer}
\end{figure}

We investigated the effect of optimizers on model accuracy using experiments 1, 2, 4 and 14. We studied the performance for four optimizers; SGD, Adam, RMSProp and Adagrad. The results showed that the Adam optimizer consistently outperformed the other optimizers across all the models. The performance metrics for these models are shown in figure \ref{fig:var_optimizer}.

\subsection{Batch size}
\begin{figure}
    \centering
    \includegraphics[scale=0.5]{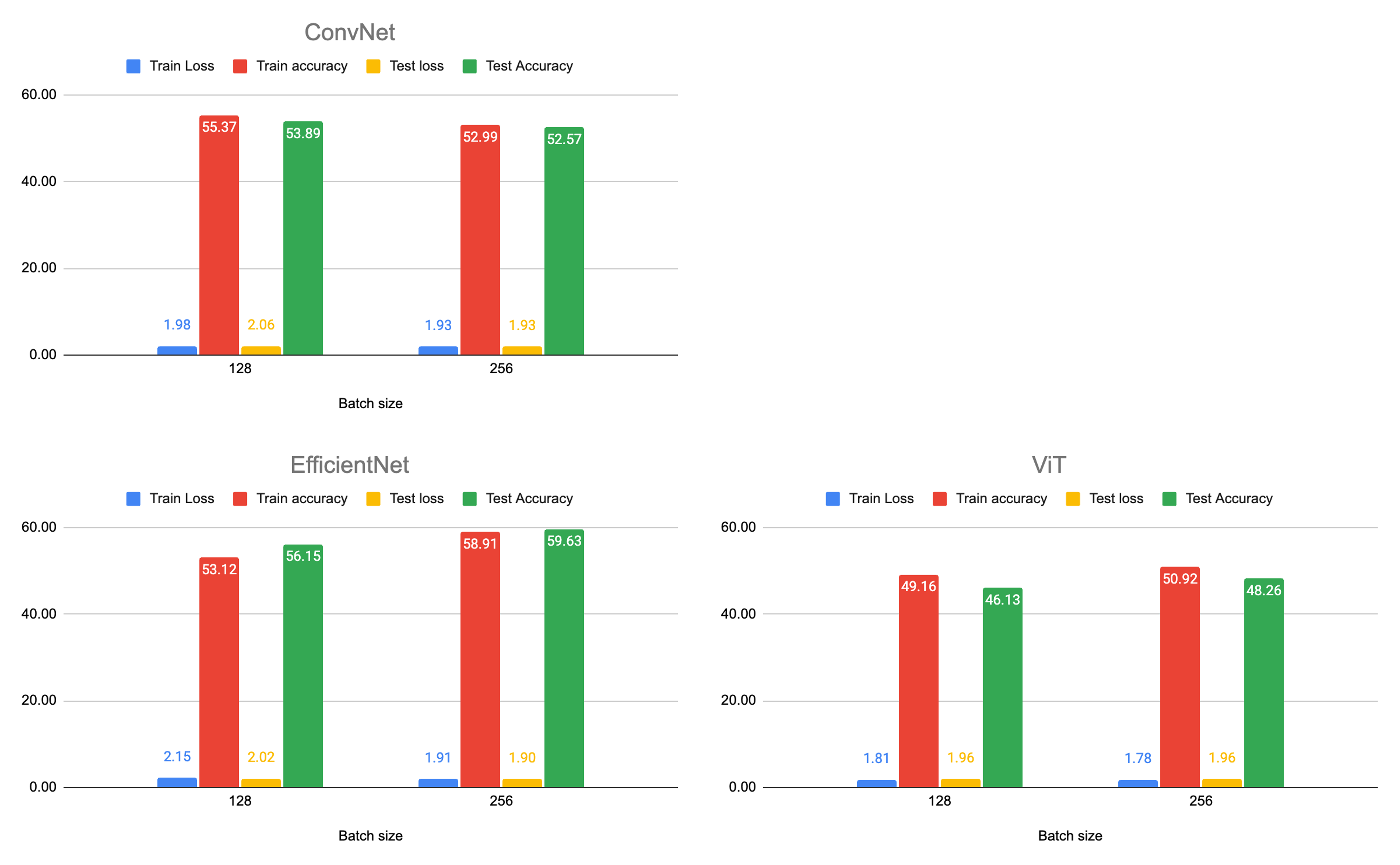}
    \caption{\small Performance of different models when batch size is 128 and 256. Both experiments implement Adam optimizer with $\epsilon=5$, $C=1$ $learning rate=1e-3$ and $epoch=100$.}
    \label{fig:var_batch}
\end{figure}

We explored the impact of batch size on model accuracy using experiments 7, 8 under one setting and 9, 10 under the other. We report the metrics of experiments 9 and 10 which uses Adam optimizer, an $\epsilon$ value of 5, and a clipping threshold ($C$) of 1. We studied the model performance for the batch size of 128 and 256. The results indicated that increasing the batch size improved the model accuracy for all the models. The performance metrics for these models are shown in figure \ref{fig:var_batch}.

\subsection{Learning rate}
\begin{figure}
    \centering
    \includegraphics[scale=0.5]{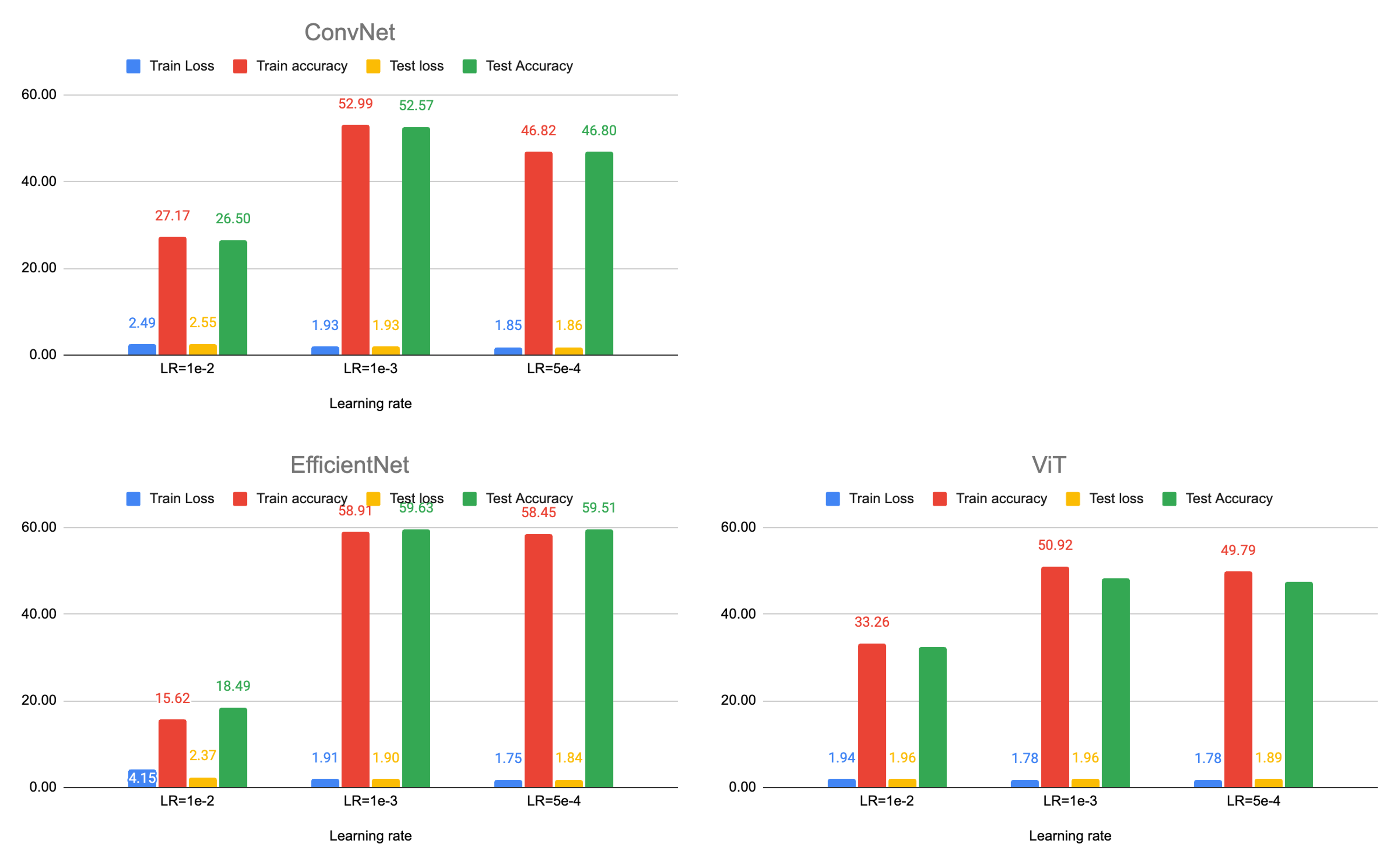}
    \caption{\small Performance of different models when learning rate is 1e-2, 1e-3 and 5e-4. All these experiments implement Adam optimizer with $\epsilon=5$, $C=1$, $batch size=256$ and $epoch=100$.}
    \label{fig:var_lr}
\end{figure}

We studied the impact of learning rate on model accuracy using experiments 10, 11 and 13. The experiments were conducted with Adam optimizer, an $\epsilon$ value of 5, and a clipping threshold ($C$) of 1. We studied the model performance for the learning rates 1e-2, 1e-3 and 5e-4. It is seen that the learning rate of 1e-2 leads to bad performance for all the models. Learning rate of 1e-3 and 5e-4 seem to have the similar performances. The performance metrics for these models are shown in figure \ref{fig:var_lr}.

\subsection{Epoch size}
\begin{figure}
    \centering
    \includegraphics[scale=0.4]{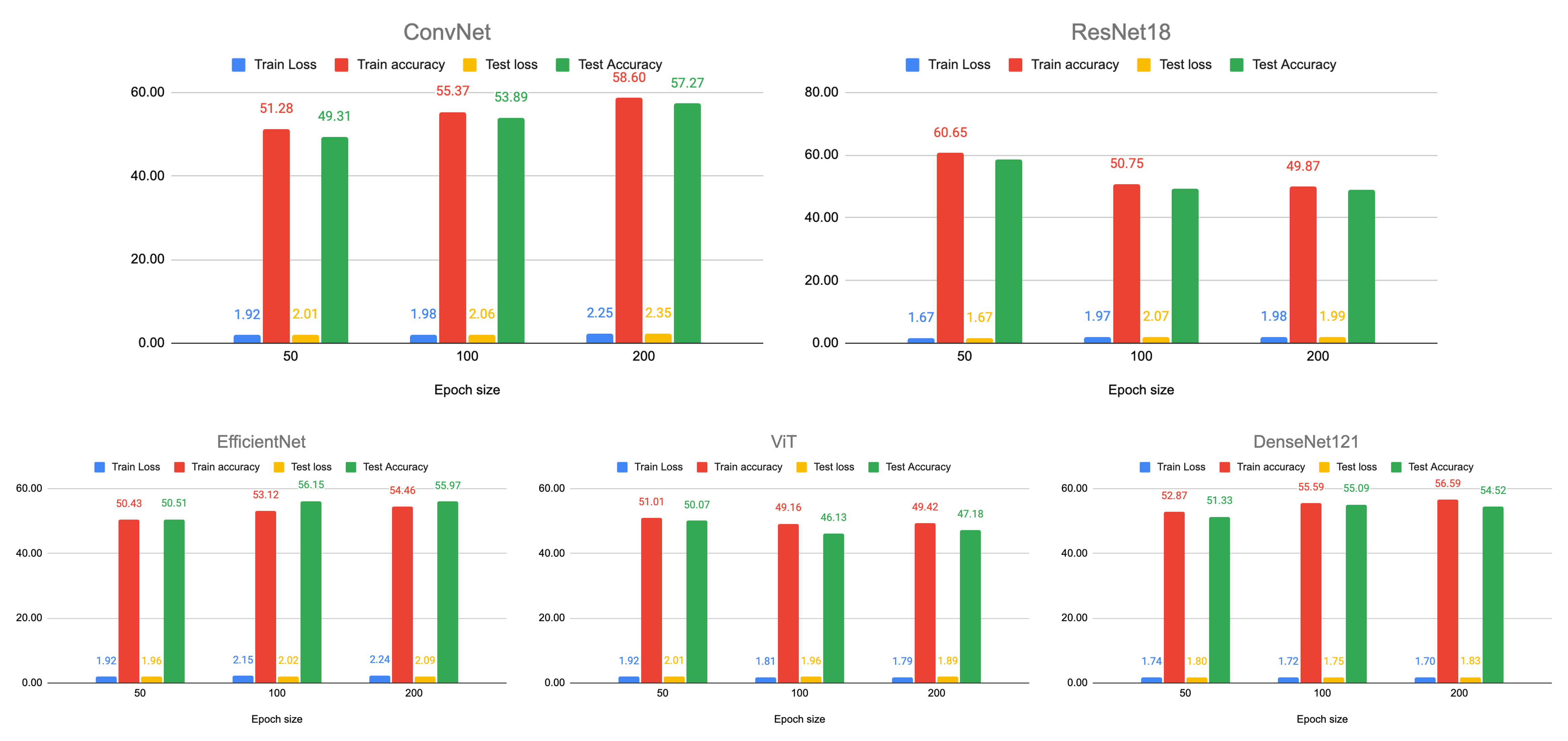}
    \caption{\small Performance of different models when learning rate is 1e-2, 1e-3 and 5e-4. All these experiments implement Adam optimizer with $\epsilon=5$, $C=1$, $batch size=128$ and $epoch=100$.}
    \label{fig:var_epoch}
\end{figure}

We studied the impact of epoch size on model accuracy using experiments 3, 9 and 20. The experiments were conducted with Adam optimizer. We studied the model performance for the epoch sizes 50, 100 and 200 keeping all other parameters constant. It is seen that the increasing the epoch size improves the performance in general, but an exception is seen for ResNet18 model. The performance metrics for these models are shown in figure \ref{fig:var_epoch}.

\subsection{Classical classifiers}
We evaluated the model accuracy of three supervised learning algorithms on the noisy data. We added a Laplace noise in the input data with $\epsilon=5.0$ and $sensitivity=1.0$. We then applied SVM, NBC and KNN on the noisy data. We tried four different kernel types in SVM algorithm. The performance metrics of the models are shown in table \ref{perf_svm} and \ref{perf_nbc_knn}.

\begin{table}
  \caption{Performance metrics of SVM algorithm on noisy data for different kernel types.}
  \label{perf_svm}
  \centering
  \begin{tabular}{ccccc}
    \toprule
        Kernel type  & Training loss   & Training accuracy & Test loss & Test accuracy \\
    \midrule
        Linear	& 0	& 100.00\%	& 0.7692	& 23.08\% \\
        Poly	& 0.4118	& 58.82\%	& 0.9615	& 3.85\%	\\
        RBF	& 0.1961	& 80.39\%	& 0.8462	& 15.38\% \\
        Sigmoid	& 0.6275	& 37.25\%	& 0.8077	& 19.23\% \\   
    \bottomrule
  \end{tabular}
\end{table}

\begin{table}
  \caption{Performance metrics of SVM algorithm on noisy data for different kernel types.}
  \label{perf_nbc_knn}
  \centering
  \begin{tabular}{ccccc}
    \toprule
    Classifier  & Training loss   & Training accuracy & Test loss & Test accuracy \\
    \midrule
        KNN Classifier (N=10)	& 0.5784	& 42.16\%	& 0.8462	& 15.38\% \\
        Naive Bayes Classifier	& 0.5784	& 42.16\%	& 0.8462	& 15.38\% \\
    \bottomrule
  \end{tabular}
\end{table}

\section{Discussion}
Our study demonstrates the importance of considering the impact of different parameters on the performance of differentially private deep learning models. It also highlights the trade-off between privacy and accuracy, which must be carefully considered when designing and implementing such models. Our findings are consistent with previous research in the field and provide valuable insights for future work in this area. Our findings can be summarized as follows.

\paragraph{Finding 1} A decrease in $\epsilon$ led to a decrease in the accuracy of all four models, indicating a trade-off between privacy and accuracy. The reason behind the decrease in accuracy with a decrease in $\epsilon$ is the trade-off between privacy and accuracy. As $\epsilon$ decreases, the level of privacy increases, but at the same time, it adds more noise to the model, which makes it difficult to learn the underlying patterns in the data, resulting in decreased accuracy. Various previous works support this observation, including the works by Abadi et al. \cite{abadi2016deep} and Shokri et al. \cite{shokri2015privacy}.

\paragraph{Finding 2} An increase in the clipping threshold (C) improved the accuracy of all four models. An increase in C improved the model accuracy for all four models because a higher clipping threshold allows for more information to be retained during the differential privacy process. This allows the model to have better accuracy while still maintaining a high level of privacy. This is consistent with the findings of previous studies on the effects of clipping on deep learning models under differential privacy settings, including work by Abadi et al. \cite{abadi2016deep}.

\paragraph{Finding 3} Our experiments have demonstrated that the Adam optimizer outperformed other optimizers. Adam optimizer is better than others because it performs well in noisy or sparse gradients, which is the case in differential privacy settings. Adam optimizer also adapts the learning rate for each parameter, which makes it more efficient in converging to the optimal solution. This is consistent with the findings of previous studies on the effects of different optimizers on deep learning models under differential privacy settings, including work by Reddi et al. \cite{reddi2020adaptive}.

\paragraph{Finding 4} Increasing the batch size improved model accuracy for all four models. Increasing the batch size improved the model accuracy for all four models because a larger batch size reduces the variance of the gradients, leading to a more stable optimization process. This is consistent with the findings of previous studies on the effects of batch size on deep learning models, such as work done by Smith et al. \cite{smith2017dont}.

\paragraph{Finding 5} Increasing learning rate helps to decrease accuracy. This finding is due to the fact that a higher learning rate causes the model to make larger updates to its weights at each iteration, potentially leading to overshooting the optimal solution. This can cause the model to converge to a suboptimal solution, resulting in decreased accuracy. This is consistent with the observations of previous studies on the effects of learning rate on deep learning models, including work by Smith et al. \cite{smith2017dont}.

\paragraph{Finding 6} Increasing epoch size helps to increase accuracy. This finding is due to the fact that more training epochs allow the model to see the training data more times, enabling it to learn more complex patterns in the data. This results in increased accuracy as the model becomes more capable of generalizing to new, unseen data. This is consistent with the findings of previous studies on the effects of epoch size on deep learning models, including work by Ajayi et. al. \cite{ajayi2021effect}.

\paragraph{Finding 7} Deep learning models are better than classical supervised learning in image classification. This finding is due to the fact that deep learning models are designed to learn hierarchical representations of data, allowing them to capture more complex patterns in the data. This is particularly useful in image classification tasks, where images contain multiple levels of abstraction. Classical supervised learning models, on the other hand, rely on manually engineered features, which can limit their ability to capture complex patterns in the data. This is consistent with the findings of previous studies on the comparative performance of deep learning models and classical supervised learning models in image classification tasks, including work by Miranda et al. \cite{miranda2022image}.

\section{Conclusion}
In summary, our study on the impact of differential privacy on image classification using five deep learning models has revealed important findings. We have observed a trade-off between privacy and accuracy, where decreasing the privacy parameter $\epsilon$ led to a decrease in model accuracy. On the other hand, increasing the clipping threshold (C), batch size, and epoch size resulted in improved accuracy for all four models. Adam optimizer outperformed other optimizers in all the experiments. Our study provides useful insights for researchers and practitioners who seek to implement differential privacy in their image classification tasks while maintaining high accuracy.

\small
\bibliography{main}

\end{document}